\documentclass[ijds,nonblindrev]{informs-ijds}

\OneAndAHalfSpacedXI
\usepackage{natbib}
 \bibpunct[, ]{(}{)}{,}{a}{}{,}%
 \def\bibfont{\small}%

\usepackage{multirow}

\TheoremsNumberedThrough     % Preferred (Theorem 1, Lemma 1, Theorem 2)
\ECRepeatTheorems

\EquationsNumberedThrough    % Default: (1), (2), ...
\MANUSCRIPTNO{}

\begin{document}
%%%%%%%%%%%%%%%%

%\setcounter{page}{1}
%\DOI{10.1287/xxxx.0000.0000}%

\RUNAUTHOR{Zhao et al.}

\RUNTITLE{Hierarchical Classification for Aviation Accident Reports}

% \TITLE{Bundling Information Goods of Decreasing Value}
\TITLE{Hierarchical Multi-label Classification for Fine-level Event Extraction from Aviation Accident Reports}

\ARTICLEAUTHORS{%
\AUTHOR{Xinyu Zhao}
\AFF{School of Computing and Augmented Intelligence, Arizona State University, Tempe, 85287, AZ, USA, \EMAIL{xzhao119@asu.edu}} %, \URL{}}
\AUTHOR{Hao Yan}
\AFF{School of Computing and Augmented Intelligence, Arizona State University, Tempe, 85287, AZ, USA, \EMAIL{HaoYan@asu.edu}}
\AUTHOR{Yongming Liu}
\AFF{School for Engineering of Matter, Transport and Energy, Arizona State University, Tempe, 85287, AZ, USA, \EMAIL{yongming.liu@asu.edu}}
} % end of the block

\ABSTRACT{%
A large volume of accident reports is recorded in the aviation domain, which greatly values improving aviation safety. To better use those reports, we need to understand the most important events or impact factors according to the accident reports. However, the increasing number of accident reports requires large efforts from domain experts to label those reports. In order to make the labeling process more efficient, many researchers have started developing algorithms to identify the underlying events from accident reports automatically. This article argues that we can identify the events more accurately by leveraging the event taxonomy. More specifically, we consider the problem a hierarchical classification task where we first identify the coarse-level information and then predict the fine-level information. We achieve this hierarchical classification process by incorporating a novel hierarchical attention module into BERT. To further utilize the information from event taxonomy, we regularize the proposed model according to the relationship and distribution among labels. The effectiveness of our framework is evaluated with the data collected by National Transportation Safety Board (NTSB). It has been shown that fine-level prediction accuracy is highly improved, and the regularization term can be beneficial to the rare event identification problem. 
% Enter your abstract
}%

%\KEYWORDS{deterministic inventory theory; infinite linear programming duality;
%  existence of optimal policies; semi-Markov decision process; cyclic schedule}

 \HISTORY{This paper is accepted in Informs Journal on Data Science on Mar 7th}

\maketitle
%%%%%%%%%%%%%%%%%%%%%%%%%%%%%%%%%%%%%%%%%%%%%%%%%%%%%%%%%%%%%%%%%%%%%%

% Samples of sectioning (and labeling) in MNSC
% NOTE: (1) \section and \subsection do NOT end with a period
%       (2) \subsubsection and lower need end punctuation
%       (3) capitalization is as shown (title style).
%
%\section{Introduction.}\label{intro} %%1.
%\subsection{Duality and the Classical EOQ Problem.}\label{class-EOQ} %% 1.1.
%\subsection{Outline.}\label{outline1} %% 1.2.
%\subsubsection{Cyclic Schedules for the General Deterministic SMDP.}
%  \label{cyclic-schedules} %% 1.2.1
%\section{Problem Description.}\label{problemdescription} %% 2.

% Text of your paper here

\section{Introduction}

In air traffic management, identifying safety risk and supporting safety improvements to mitigate risk is of great importance for the next generation of national air transportation systems. The National Transportation Safety Board (NTSB) has investigated and collected more than 60000 aviation accident and incident reports with labeled sequences of accident events. 
It is of great importance to improve aviation safety by preventing repeat accidents. The engineers at NTSB have spent a lot of time recording past aviation accidents and understanding the crucial causes of past accidents. Since the accident is usually recorded in raw text format, the root cause analysis requires experts with domain understanding. However, as more and more accident data is collected, such an analysis strategy becomes very inefficient. 

{With the development of Natural Language Processing (NLP) techniques, text mining methods have become popular in traffic management. For example, \citet{yao2021twitter} proposes to analyze Twitter messages using text mining to predict the traffic on the next day. \citet{rath2022worldwide} proposes to utilize the state-of-the-art NLP model to use Wikipedia data to predict the typology of city transport. More specifically, people started to design more efficient text mining tools to automatically identify all accident events from the accident report using historical accident report data with identified critical events. When new accident reports are available, the underlying events can be automatically identified using powerful supervised learning methods. There are many existing works on the analysis of incident reports. For example, \citet{pereira2013text} proposes using traffic incident record messages to predict the duration of the incident. The knowledge graph is also used to reveal the relationship among critical components in ship collision accidents \citep{gan2023knowledge}. Many researchers have started bringing NLP into the aviation domain. \citet{yelundur2016event} proposed to apply Bayesian logistic regression to classify four major events, namely loss of separation, deviation and ATC anomalies, Ground and landing events, and Loss of control. A question and answer system is developed for the aviation domain with graph knowledge \citep{agarwal2022knowledge}. \cite{abedin2010cause} was applying weakly supervised semantic lexicon construction to identify causes from aviation incident reports. Similarly, \cite{robinson2018multi} focus on cause identification task as well from a multi-label classification perspective. 
However, due to the limited capacity of these traditional machine learning models, they were unable to utilize a large number of accident reports and were unable to model the contextual relationship between words.}

{In recent years, deep learning models have been proposed to analyze the data from the aviation accident report. \citet{dong2021identifying} proposed an attention-based long short-term memory (LSTM) to identify the six most frequent impact factors from incident reports recorded by the Aviation Safety Reporting System (ASRS). Another work based on ASRS proposed identifying the 14 most important cause types, such as physical factors,  physical environment, proficiency, etc. These works have achieved satisfactory results on a limited number of factors \citep{abedin2010cause}. \citet{zhao2022event} have proposed another method to identify 58 main causes in the NTSB accident report using the LSTM model. Recently, Bidirectional Encoder Representations from Transformers (BERT) model has been proposed in the Natural Language Processing domain. The BERT model has also been applied to accident report modeling. For example, a BERT-based question-answering system has been proposed based on the ASRS dataset \citep{kierszbaum2020applying}. However, the major limitation of existing NLP models on the aviation accident report data is that we can only identify coarse categories from reports.} Existing works still require experts' effort to specify the details within those categories. For example, it is very challenging for algorithms to tell what the exact physical factors of accidents are, which is an important task in aviation accident modeling. {\cite{zhang2020bayesian} applied the Bayesian network to the NTSB dataset to study the causal and dependent relationships among a wide variety of detailed contributory factors. \cite{rao2020state} mentioned that there exists inconsistent coding and short chain lengths associated with general aviation accidents in NTSB. To be specific, a great variety of chains (152 different occurrence chains), with more $82\%$ of the accidents having an average chain length of two or less. By consistently identifying fine categories from reports, we will compensate for the missing logic in an accident event chain. Finally, \cite{tanguy2016natural} also provides evidence that the existing taxonomy categories are generally too broad to identify a specific characteristic of an event in ASRS. }  

\begin{figure}[h!]
\centering
\includegraphics[width=0.85\textwidth]{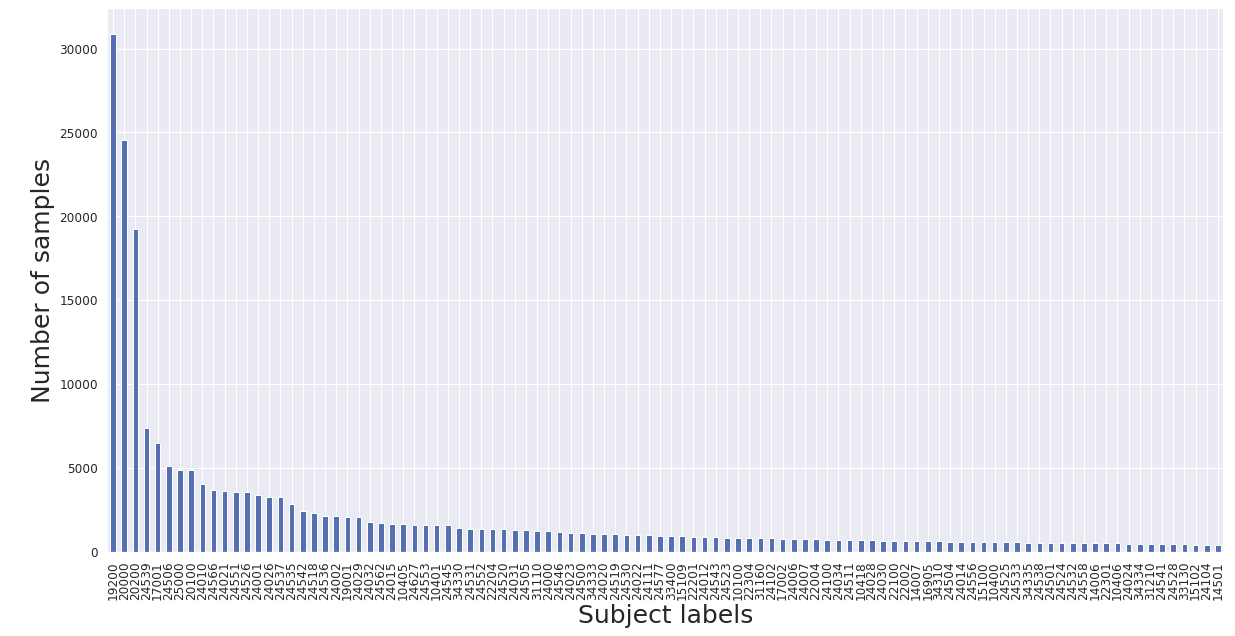}
\caption{Label statistics for 100 most frequent fine-level labels in NTSB}
\label{distribution}
\end{figure}

There are four major challenges if we want to identify the fine categories from the accident reports. We will use accident reports from the National Transportation Safety Board (NTSB) as an example to illustrate the challenges as follows:
1) \textbf{Multiple accident events for each accident report}: in general, there can be multiple accident events corresponding to each accident report. The number of accident events in each report is also typically unknown.
{2) \textbf{Complex correspondent}: Many occurrence codes or even subject codes correspond to multiple sentences from the original aviation accident report. For example, the occurrence code "Airplane/component/system failure/malfunction" corresponds to multiple sentences in the original aviation accident report, including "loss of the alternator", "loss of the cockpit lighting", loss of compass", etc. These keywords are never directly mentioned in the sentence but rather inferred from several sentences or the entire manuscript.}
3) \textbf{Lack of data for some fine-level categories}: when we are trying to build a supervised learning model, we must ensure enough samples for all categories. For example, according to the event taxonomy defined by NTSB, there are 58 occurrence codes in the coarse-level labels and 1432 subject codes in the fine-level labels. In this case, since the coarse category is always a general abstract of many possible causes, it often accumulates enough samples for us to build a supervised learning model. However, as a subset of coarse categories, many fine categories might suffer a few-shot learning problem \citep{wang2020generalizing}. For example, there are 1191 fine categories that have less than 100 accident reports, and 602 fine categories have less than 10 samples in the NTSB data. 
4) \textbf{Data imbalance issue}: as the number of fine categories is much larger than the coarse categories, it resembles the natural non-uniform distribution in real-world scenarios. The fine categories appear to follow a long-tail distribution pattern, as shown in Fig. \ref{distribution}. This phenomenon makes the model tend to be over-confident in the frequent categories and hard to recognize the rare categories. The two challenges mentioned here are commonly seen when dealing with a dataset with a large label size. In this paper, we propose to deal with the problems by utilizing the category taxonomy and view this problem as a hierarchical classification task.

\begin{figure}[h!]
\centering
\includegraphics[width=0.8\textwidth]{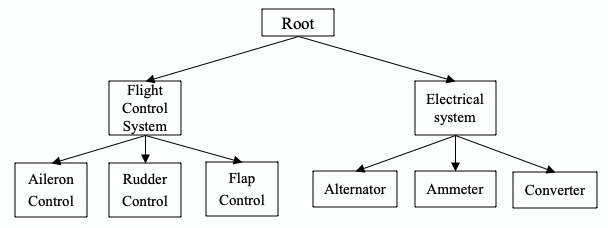}
\caption{An example of label taxonomy defined by NTSB}
\label{example}
\end{figure}

{In the NTSB accident report, the accident events may consist of a hierarchical structure. Here, we define a more general definition of label hierarchy using a directed acyclic graph (DAG), where each node may have more than one parent node.} The parent nodes represent a high-level abstract of all its children. Fig. \ref{example} presents an example of label taxonomy in the aviation domain defined by NTSB. The figure shows that flight control systems and electrical systems are coarse categories. They provide general information about what went wrong during the accident. To classify the fine-level category, such as flap control system or rudder control system, since the sample size for coarse-level categories is much larger than for fine-level categories, it is much easier to discriminate between the flight control system and the electrical system. On the other hand, it is challenging for the model to directly recognize aileron control or rudder control due to the limited sample size. To address this challenge, we propose to bring the hierarchical information from the labels into the model instead of treating each category level independently. 

In the literature, bringing the label hierarchy to the classification problem is often known as hierarchical classification, which has shown the capacity to improve fine-level prediction accuracy in the case of unbalanced data and small sample sizes for some categories.
The hierarchical classification approaches are typically grouped into two directions: (1) Local approaches train multiple classifiers according to the label taxonomy, and the prediction follows a top-down manner. (2) Global approaches introduce a single classifier that can handle different levels of categories together \citep{silla2011survey}. The local approaches train category-wise models or level-wise models. Level-wise models follow a top-down manner where coarse categories are predicted first, and the fine level is predicted using the inferred coarse-level information as additional predictors. It is much easier to capture local information from the data. However, such an approach often suffers from the error propagation problem as fine level prediction requires very accurate coarse-wise models. Global approaches treat the problem as a whole which might not capture the local information very well. However, it usually utilizes the hierarchical information much better by simultaneously optimizing for both coarse categories and fine categories together. For more detailed literature on general hierarchical classification, please refer to Section \ref{sec: literature review}.

\begin{figure*}[h]
\centering
\includegraphics[width=1\textwidth]{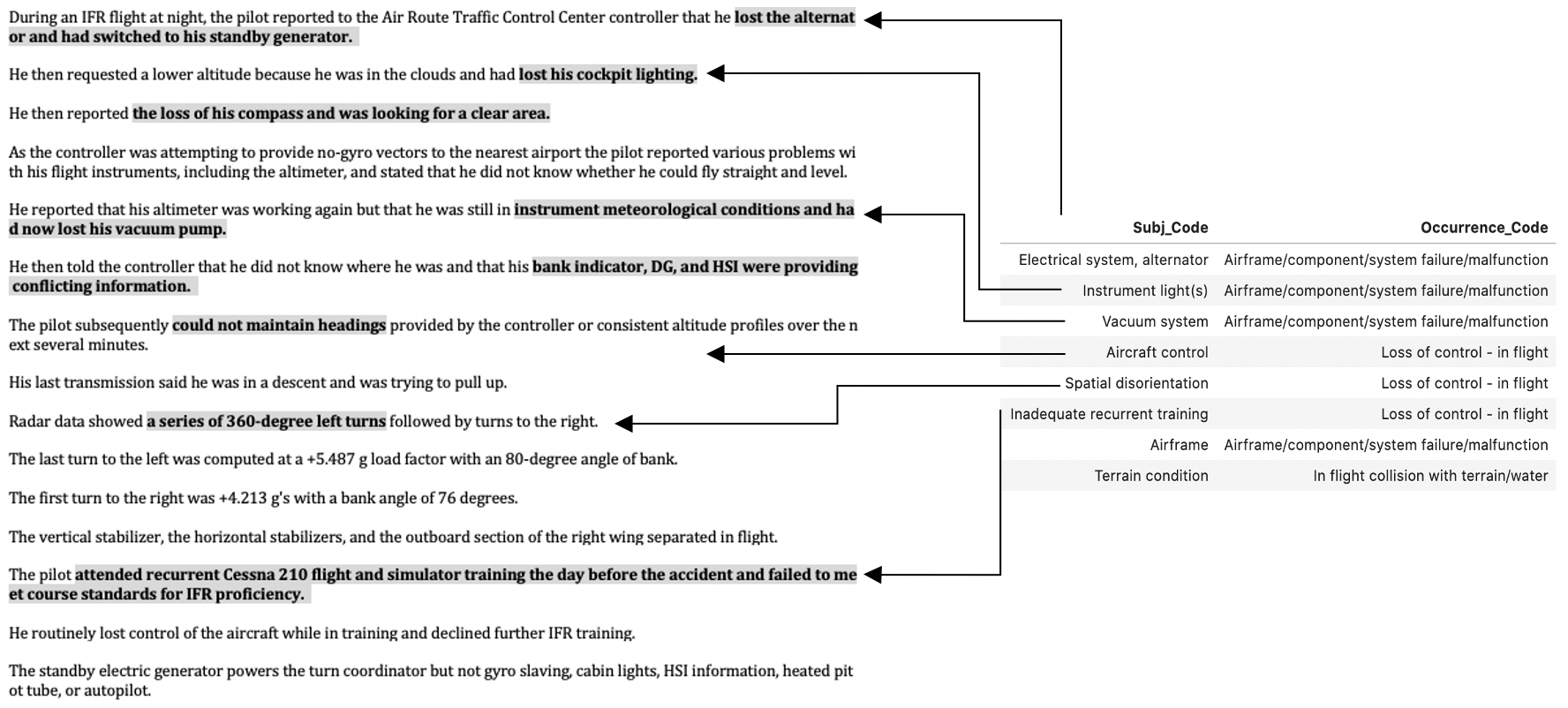}
\caption{Aviation accident report to event labels: Report 20001208X07734 from NTSB. The left side is the raw accident report. On the right side is the event sequence labeled by NTSB. We highlight the keywords in the narrative reports and plot their relationship with the corresponding event labels}
\label{example_NTSB}
\end{figure*}

In conclusion, we propose to combine the hierarchical multi-label classification approaches with a state-of-the-art BERT model to identify multiple events with fine-level categories from aviation accident reports. 
Our experiments are conducted based on the reports collected by NTSB. As shown in Fig. \ref{example_NTSB}, the accident report is associated with multiple event labels. The occurrence code represents a coarse-level description of the accident, and the subject code represents a fine-level description. Given that the size of subject labels is too large, the traditional multi-label classification algorithm cannot achieve satisfying results due to data sparsity and data imbalance. This work aims to improve the fine-level classification by incorporating the event taxonomy into the hierarchical classification. The proposed method takes advantage of the state-of-the-art BERT model to handle the raw text data with the complex contextual word dependency. Furthermore, related to the hierarchical multi-label classification, the proposed method takes advantage of both the local and global approaches for hierarchical classification by integrating hierarchical regularization, hierarchical attention, and hierarchical label distribution penalty to extract multiple events simultaneously from the accident report.

In summary, the major contributions of this paper are listed as follows:
\begin{enumerate}
    \item We formulate the cause identification of aviation accident reports by combining the hierarchical classification and the state-of-the-art BERT model. By comparing with the BERT-based flat classification model, we demonstrate that the hierarchical information in the event taxonomy can benefit the fine-level classification accuracy. 
    \item We propose a novel hierarchical classification framework that benefits from both the local and global approaches. More specifically, our proposed hierarchical classification approach consists of three major components: 
    \begin{enumerate}
    \item Recursive regularization: Recursive regularization is added to encourage the model parameter to be similar for the sibling nodes to the parent nodes.
    \item Hierarchical attention: Hierarchical attention is added to guide the attention module of the fine-level model using the coarse-level information. 
    \item Hierarchical Label distribution penalty: We propose a novel hierarchical label distribution penalty term. The component can further improve the classification accuracy on rare labels by dealing with the overconfidence issue on frequent labels. 
    \end{enumerate}
    \item We have applied the proposed method to the NTSB accident report data and have demonstrated the improvement of the multi-label classification accuracy, especially for rare events.

\end{enumerate}

The paper is organized as follows. We first review existing works on texting mining on narrative reports and state-of-art hierarchical classification methods in the section. \ref{sec: literature review}. We further introduce some basic notations and the proposed methods in Section \ref{sec: Methodology}. The experiment results will be discussed in the next Section \ref{sec: Experiment Setup}. Moreover, we finally conclude the paper in Section \ref{sec: Conclusion}.
\section{Literature Review  \label{sec: literature review}}
We would like to review the related literature in both the areas of accident report analysis and hierarchical classification. More specifically, from the application perspective, we will first give a brief overview of existing works on automating the information extraction for narrative reports in Section \ref{sec: Information retrieval for narrative reports}. From a methodology perspective, we will review the recent progress on hierarchical classification in Section \ref{sec: Hierarchical classification}. 

\subsection{Information retrieval for narrative reports  \label{sec: Information retrieval for narrative reports}}
It is necessary to provide an efficient prevention strategy when an accident happens by analyzing past accidents. Therefore, more researchers started developing tools to automatically extract information from narrative reports. The fundamental of the algorithms is the meaningful event taxonomy or impact factors defined by domain experts. There are various types of taxonomic structures defined early in the aviation domain. Three codifications are designed specifically for ASRS, which are the ASRS codification, the X-Form, and the Cinq-Demi \citep{ferryman2006happened}. The ASRS codification is currently used to describe the accident in the ASRS database, which presents an abstract of the accident through the "Chains of Events" and "Human performance consideration". There are nine coarse-level categories from the ASRS codification: Aircraft, Events, Maintenance Factors, Etc. Under each course category, multiple fine categories are included. For example, around 180 subcategories are under the Aircraft category, and around 130 subcategories are under the Event category.
Another widely used accident reporting system is NTSB. It designed three-level categories to describe the accident: Phase, Occurrence, and Subject. There are around 56 Phase categories, 58 Occurrence categories, and 1432 Subject categories. Based on those well-defined taxonomies, many researchers have started designing efficient information extraction frameworks to automate the accident reporting system in the aviation domain. \citet{huang2019hierarchical} developed algorithms to extract the information from the ASRS dataset according to the X-Form taxonomy. They proposed a template-based framework to identify the related keywords in the reports, and 20 ASRS reports were selected for validation purposes. The potential issue with this method is to develop an exhaustive set of keywords and expressions. \citet{abedin2010cause} proposed an automatic keyword extraction strategy via semi-supervised semantic lexicon learning and validated the efficiency of the model with all 140599 reports from 1998 to 2007. To further improve the classification accuracy, many advanced machine learning methods are applied, such as Naive Bayes \citep{shi2017data}, Bayesian logistic regression \citep{yelundur2016event}, and latent semantic analysis \citep{robinson2018multi}. It is worth noting that all the previous works are only conducted on small label sizes. As mentioned by \citet{tanguy2016natural}, fine-grained investigation of sources of danger is very challenging. It is very difficult for the experts to tell the difference between two closely related values at a detailed level. Existing works can only be implemented over the coarse level identification. In this work, we propose a framework trying to identify the fine-level categories by using the category taxonomy. 

\subsection{Hierarchical classification  \label{sec: Hierarchical classification}}
In many real-world applications, we can see the hierarchical category structure. For example, in the e-commerce system, products are classified into categories and subcategories \citep{karamanolakis2020txtract}. In the healthcare system, the taxonomy of patients' symptoms helps the doctors diagnose much easier \citep{ruggero2019integrating}. Motivated by those hierarchical representations, researchers start incorporating knowledge from the hierarchical taxonomy into traditional classification methods. Most existing hierarchical classification methods fall into two categories which are local approaches and global approaches \citep{silla2011survey}. Local approaches focus on building independent classifiers for each category or level from the taxonomy. It aims at extracting the most comprehensive local information from the samples. They usually follow a top-down prediction strategy according to the taxonomy \citep{koller1997hierarchically,costa2007comparing,kiritchenko2004hierarchical}. One major problem of the local approaches is the error-propagation and the expensive computational cost. Since multiple classifiers are developed independently, the model parameters increase rapidly, and the bias from different classifiers will accumulate. Different from local approaches, global approaches treat the problem as a whole. It builds a single model and predicts categories from different levels together \citep{cerri2012genetic,cai2004hierarchical,gopal2013recursive}. However, global approaches might not capture the local information very well. They require fewer training resources and are more sensitive about the relationship across different levels among all categories. Considering all the pros and cons of the two methods, recent works have started building hybrid methods to take advantage of both approaches. \citet{huang2019hierarchical} proposed a hybrid approach through a recurrent neural network. The output of each layer corresponds to a level from the category taxonomy. By adding a global output layer as the final layer of the network, the proposed method optimizes local loss and global loss simultaneously. To further enhance the relationship between the input and labels, a hierarchical attention-based approach is proposed, which also fall into the category of a hybrid approach. By leveraging the hierarchical structure, the methods propagate the attention weight top-down. Inspired by the aforementioned methods, we proposed a novel hierarchical attention-based framework. 

Unlike previous work, since our task suffers a long-tail distribution problem, we need to find an efficient solution to deal with the data sparsity and imbalance challenge. Recent studies have shown that we can leverage the hierarchical structure to deal with those problems. A new sampling strategy is proposed to deal with the imbalanced hierarchical dataset \citep{pereira2021toward}. \citet{wu2020solving} proposed a rejection strategy based on the label taxonomy to identify uncertain examples. Other works focusing on the few-shot learning settings have also proved the value of label hierarchy. Considering the relationship among siblings belonging to the same level of the category, \citet{xu2019hierarchical} introduced Label Distribution Learning to solve the small training set issue. \citet{li2019large} proposed a novel hierarchical feature extractor that achieves state-of-the-art results for a few-shot learning task. Another two-stage few-shot learning approach is proposed in \citep{liu2020many}. It argued that as the size of fine-level labels grows dramatically, the limited sample size for each category can hardly provide enough information for existing models to distinguish those. However, the sample size from the coarse-level categories is usually sufficient for a reliable supervised model. The paper develops a memory-augmented hierarchical classification network based on those observations. Multi-layer perceptron and K-nearest-neighbor are developed for coarse prediction and fine prediction, respectively, due to their sample size. Following this line of research, we also design a two-stage approach and feed the reliable coarse-level predictions into our fine-level model for fine-tuning purposes. We guide the model training process with a recursive regularization term and a label distribution penalty to improve the prediction accuracy on few-shot categories. 
% text mining for accident reports
% hierchical classification 
\section{Methodology  \label{sec: Methodology}}
In this section, we will introduce the proposed model. More specifically, we will first review the related methods in Section \ref{sec: Preliminary}. We will then introduce the proposed hierarchical multi-label classification BERT model in Section \ref{sec: proposed method}. 

\subsection{Review of related methodology \label{sec: Preliminary}}
In this subsection, we will review the three related methods in handling the hierarchical labels, including the hierarchical attention model in Section \ref{sec: Hierarchical Attention}, recursive regularization in Section \ref{sec: Recursive Regularization}, and label distribution learning in Section \ref{sec: Label Distribution Learning}. 

\subsubsection{Review of Hierarchical Attention \label{sec: Hierarchical Attention}}
The attention mechanism is widely used in recent deep-learning approaches. It helps us build the connection between labels and input. By adding the attention module, we can extract the most relevant component of the input to the label \citep{vaswani2017attention}. To leverage the label hierarchy information through the attention module, hierarchical attention is proposed by \cite{huang2019hierarchical}. With the assumption that coarse-level information can help us narrow down the relevant component for fine-level prediction, we can propagate coarse-level attention to fine-level by building better semantic representation. Let $S^h\in\mathbb{R}^{|C^h|\times d_a}$ denote the embedding vector for $h^{th}$ level of the category. $|C^h|$ represents the label size, and $d_a$ is a hyperparameter for the embedding dimension. The attention weight for $h^{th}$ category level can be calculated as 
\begin{equation}
    W_{att}^h = \mathrm{Softmax}(S^h \cdot O_h),\quad O_h = \tanh (W_s^h\cdot V_h^T), \label{eq: att_weight}
\end{equation}
where $V_h \in\mathbb{R}^{N\times 2u}$ is the semantic representation of the input, and $N$ is the length of the input. We further send $V_h$ through a fully connected layer with weight matrix $W_s^h\in\mathbb{R}^{d_a\times 2u}$ to get $O_h \in\mathbb{R}^{d_a\times N}$. After getting the attention weight, we will further calculate the weighted text-category attention matrix $K^h\in \mathbb{R}^{|C^h|\times N}$ with the local prediction on $h^{th}$ level of the category $P^h_L$
\begin{equation}
    K^h = \mathrm{Broadcast}(P_L^h)\otimes W_{att}^h \label{eq: weighted_attention}
\end{equation}
Here, the $\mathrm{Broadcast}$ operation is to make the shape of $P^h_L$ compatible with attention matrix $W_{att}^h$ so that we can conduct the element-wise multiplication $\otimes$. $K_h$ is further averaged along the category dimension, and then we can get the vector $\Tilde{K}^h\in \mathbb{R}^{N}$ that shows the most relevant component from our input to the $h^{th}$ category level. Finally, the updated semantic representation can be calculated with the weight attention vector $\Tilde{K}^h$
\begin{equation}
V_{h+1} = \omega^{h} \otimes \mathbf{V},\quad \omega^{h} = \mathrm{Broadcast}(\Tilde{K}^h) \label{eq: weight_representation}   
\end{equation}

It is worth noting that the attention weight is propagating to the lower level through constructing a new semantic representation. In our framework, we simplify the calculation and let coarse-level attention directly influence the attention weight on the fine level. 
\subsubsection{Review of Recursive Regularization  \label{sec: Recursive Regularization}}
Recursive regularization is first proposed by \citet{gopal2013recursive}. as a global strategy to deal with hierarchical classification. It encourages the model parameters among sibling nodes in the label hierarchy to be similar. Later work extends this method to a deep learning setting \citep{peng2018large}. Given the observation that coarse-level categories usually have more training samples than fine-level categories. It is easier to get the optimal parameters for the coarse category. Thus, regularizing the children's nodes to have similar parameters to their parent will help improve the fine-level classification even though we have fewer training samples on fine categories. To be formal, let $w_i$ be the parameters in the final fully-connected layer for all categories. $l_i^{j}$ represents the children of $l_i$. Then, the recursive regularization term can be added to the loss function as shown in Eq. (\ref{eq: recursive regularization}). 
\begin{equation}
    \lambda(\mathbf{W}) = \sum\limits_{l_i\in L}\sum\limits_{l_i^{(j)}}\frac{1}{2}\|w_{l_i}-w_{l_i^{j}}\|^2 \label{eq: recursive regularization}
\end{equation}
Here, the final loss function will be $H+C\lambda(\mathbf{W})$, where $H$ is the cross-entropy loss. Inspired by this formulation, our model also implements a similar regularization idea. However, since we mainly express the hierarchical information through the attention module, we will add this penalty by embedding vectors from different levels of categories.
\subsubsection{Review of Label Distribution Learning  \label{sec: Label Distribution Learning}}
Another algorithm that can deal with the small training set issue through the label hierarchy is called label distribution learning \citep{xu2019hierarchical, geng2016label}. It is argued in the paper that the fine classifiers tend to overfit due to the small training size. We can leverage the relationship among siblings in the label hierarchy to get additional supervision information. As mentioned in the paper, Label Distribution Learning is a more general learning framework for single-label learning or multi-label learning. Under the label distribution learning setting, each instance $x_i$ is associated with a label distribution $D_i=\{d_{x_i}^{y_1}, d_{x_i}^{y_w}, \dots,d_{x_i}^{y_n}\}$ where $y_j$ is the $j^{th}$ label. $d_{x_i}^{y_j}\in[0,1]$ represents the degree of $y_j$ to instance $x_i$ where $\sum_j d_{x_i}^{y_j}=1$. $d_{x_i}^{y_j}$ can also be represented as the form of conditional probability where $d_{x_i}^{y_j} = P(y_j|x_i)$. 

To be more formal, Let $\mathcal{X}\in\mathbb{R}^q$ denotes the input space and $\mathcal{Y}=\{y_1, y_2,\dots,y_c\}$ denotes the complete set of labels. Given a training set $S=\{(x_1,D_1),(x_2,D_2),\dots,(x_n,D_n)\}$ Then the goal of the Label Distribution Learning becomes learn a conditional probability mass function $p(y|x)$ from S, where $x\in \mathcal{X}\ and\ y\in\mathcal{Y}$. Let the parameter $\theta$ to control the conditional probability $p(y|x, \theta)$. Then we need to find the optimal parameter set to generate similar distribution to $D_i$ given the instance $x$. Multiple criteria can be used to evaluate the difference. Kullback-Leibler (KL) divergence is a frequently used measurement to describe the similarity between two distributions. Moreover, the optimization objective can be formulated as follows:
\begin{equation}
    \theta^{*}=\arg\min\limits_{\theta}\sum\limits_{i}\sum\limits_{j}(d_{x_i}^{y_j}\ln\frac{d_{x_i}^{y_j}}{p(y_j|x_i,\theta)}). \label{label distribution learning}
\end{equation}
Under the hierarchical classification setting, the true label distribution $D$ is usually not achievable. Moreover, there are multiple ways to construct meaningful label distribution according to the label hierarchy. One possible way to get the label distribution mentioned by \cite{xu2019hierarchical} is through leveraging the knowledge of the Number of Common Nodes with the True Path (NCNTP). We propose a novel way to achieve label distribution in our work. We propose to use the Bayes rule to get the fine-level distribution through our prediction of the coarse-level distribution. 
\subsection{Proposed Model for Hierarchical Multi-label classification \label{sec: proposed method}}
In this section, we will introduce the proposed Proposed Model for the Hierarchical Multi-label classification model and combine it with the BERT model to analyze the accident report data. Section \ref{sec: Problem definition} presents the overall problem definition. Section \ref{sec: Model Architecture} presents the overall model architecture. 
Section \ref{sec: coarse-level model} presents the architecture of coarse-level model. The three major components of the fine-level model are introduced in Section \ref{sec: proposed Hierarchical attention}, \ref{sec: Recursive regularization proposed}, and \ref{Label distribution penalty-proposed}, respectively. Finally, the loss function that integrates all the components together is introduced in Section \ref{sec: loss function}.

\subsubsection{Problem definition  \label{sec: Problem definition}}
This paper aims to design an information extraction framework for aviation accident reports. As shown in Fig.\ref{example_NTSB}, there is a two-level description of the reports, which are called Occurrence and Subject. Occurrence is a coarse level event description containing 56 categories, while Subject is a fine level description with 1432 categories. Motivated by the label hierarchy, we want to build a coarse-to-fine hierarchical classification model to address the major challenges of the lack of enough samples in many fine-level categories. 

Under the supervised learning setting, our problem can be formulated as follows: Given a set of training samples $\mathcal{D}=\{(\boldsymbol{x}_n, \boldsymbol{y}_n, \boldsymbol{z}_n)\}_{n=1}^N\subseteq \mathcal{X}\times \mathcal{Y}\times \mathcal{Z}$. Since our input is accident reports, each $\boldsymbol{x}_n$ represents the input documents and $\mathcal{X}$ denotes the space of the natural language text where $\mathcal{X}=\{\omega_1, \omega_2,\dots,\omega_M\}$ can be viewed as a sequence of words. $\boldsymbol{y}_n=[y_1, y_2, \cdots, y_{L_1}]\in \mathcal{Y}=\{0,1\}^{L_1}$ denotes $L_1$ number of coarse-level categories. $\boldsymbol{z}_n=[z_1, z_2, \cdots, z_{L_2}]\in \mathcal{Z}=\{0,1\}^{L_2}$ denotes $L_2$ number of fine-level categories. Our goal is to get an accurate fine-level prediction based on our observations. 
\begin{equation}
    \min\limits_{\theta}\mathbb{E}_{\{\boldsymbol{x}, \boldsymbol{y}, \boldsymbol{z}\}\sim \mathcal{D}} (-\log \mathbf{Pr}(\boldsymbol{z}|\boldsymbol{x};\theta)) \label{eq: loss}
\end{equation}
where $\theta$ denotes the model parameters. Based on the observation that coarse level categories usually have enough samples for the training purpose. In order to further improve the fine level prediction, the objective function can be broken into two components as follows: 
\begin{equation}
    \mathbf{Pr}(\boldsymbol{z}|\boldsymbol{x};\theta) = \sum\limits_{\boldsymbol{y}\in  \mathcal{Y}} \mathbf{Pr}(\boldsymbol{z}|\boldsymbol{x},\boldsymbol{y};\theta_f)\mathbf{Pr}(\boldsymbol{y}|\boldsymbol{x};\theta_c), \label{eq: fine coarse}
\end{equation}
where $\theta_f$ and $\theta_c$ represent the fine-level model parameters and coarse-level model parameters, respectively. Thus, the problem can be transformed into a two-stage decision framework. Firstly, we need to design a coarse-level classifier. Then we need to design a fine-level classifier that can utilize the information predicted from the coarse model. It is worth noting that the proposed procedure is similar to the top-down prediction framework of the local approaches in hierarchical classification. To mitigate the error propagation issue, the information from the coarse models is not added to fine models directly as traditional local hierarchical classification approaches \citep{koller1997hierarchically}. We build a hierarchical attention module to get better fine-level feature representation with the guidance of coarse-level information. We further optimized embedding vectors for coarse categories and fine categories together following a global approach framework. 
\begin{figure}[h!]
\centering
\includegraphics[width=1\textwidth]{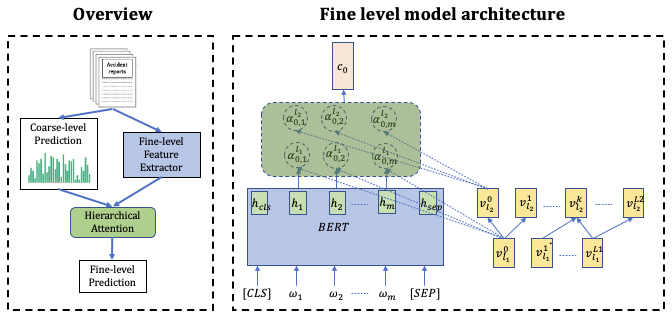}
\caption{Overview of the two-stage hierarchical attention.}
\label{model}
\end{figure}
\subsubsection{Overview of the model architecture  \label{sec: Model Architecture}}
Figure \ref{model} presents an overview of the proposed framework. There are three major components of the framework: a coarse-level classifier, a fine-level feature extractor, and a hierarchical attention module. Our methods provide coarse-level prediction through fine-tuning on a pre-trained BERT model \citep{DevlinCLT19}. BERT (Bidirectional Encoder Representation from Transformers) is one of the most powerful pre-trained deep language models that have achieved the state of the art results in many text mining tasks. A BERT-based classifier can provide satisfying results with a sufficient amount of coarse data. Therefore, it can further guide our fine classifiers. We first construct the deep representation for the accident reports with BERT for fine-level prediction. Then, the text features will feed into the hierarchical attention module and the coarse category distribution and make the fine-level prediction. Finally, we develop two penalty terms in our loss function according to the label distribution and label hierarchy to deal with the long-tail distribution over fine categories. In the following sections, we will mainly introduce the major components of the fine-level classifier.

\subsubsection{BERT-based feature extractor  \label{sec: feature extractor}}
In order to get meaningful text representation, we apply the BERT model to encode text into embedding space to extract the text embedding for future use. As shown in Figure \ref{model}, the input document  is a sequence of M words $\boldsymbol{x}=[\omega_1, \omega_2,\dots,\omega_M]$. We apply the BERT model to get the semantic representation of the text data. The discrete word $\omega_i$ are first mapped into Key $\boldsymbol{k_i}$, Value $\boldsymbol{v_i}$ and Query $\boldsymbol{q_i}$ according to the self-attention mechanism. The contextual representation of the word $\boldsymbol{z_i}$ is computed as follows
\begin{equation}
    \boldsymbol{z_i} = \mathrm{Softmax}(\frac{\boldsymbol{q_i}\boldsymbol{k}}{\sqrt{d_k}})\boldsymbol{v} \label{eq: contextual}
\end{equation}
The BERT model is built upon multiple self-attention layers and feed-forward connections. In the last layer, we can get the semantic representation of each word based on its correlation to all other words in the document. Here we will use $\boldsymbol{H}=[\boldsymbol{h}_1, \boldsymbol{h}_2,\dots,\boldsymbol{h}_M]\in \mathbb{R}^{M\times u}$ to represent the final embedding vector calculated by BERT where $u$ is the embedding dimension. To further regularize the overfitting issue, a drop-out layer is added to the final layer of the BERT model. 

\begin{figure}[h!]
\centering
\includegraphics[width=0.6\textwidth]{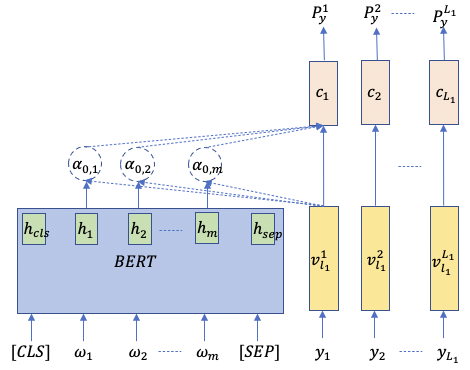}
\caption{Coarse level model architecture}
\label{coarse_model}
\end{figure}

\subsubsection{Coarse-level model  \label{sec: coarse-level model}}
We will then train a coarse-level multi-label classification model based on the extracted features from the BERT-based feature extractor to be used as the local approach to guide the fine-level multi-label classification. The coarse-level prediction model consists of the following components: a BERT-based feature extractor, an attention module for identifying the coarse-level relevance score, and the final BCE loss function for multi-label classification. Figure. \ref{coarse_model} illustrates the architecture of the coarse-level model. Let $V_{l_1}=[\boldsymbol{v_{l_1}^1},\boldsymbol{v_{l_1}^2},\cdots,\boldsymbol{v_{l_1}^{L_1}}]\in\mathbb{R}^{L_1\times u}$ denote the coarse label embedding. We first calculate the relevance score between documents and labels as follows:
\begin{equation}
    e_{ij} = \boldsymbol{v_{l_1}^i}\cdot \boldsymbol{h^j} \label{eq: relevance}, 
\end{equation}
where $e_{ij}$ is the relevance score between $i^{th}$ label and $j^{th}$ word in the document. The attention weight can be further calculated through a Softmax function.
\begin{equation}
    \alpha_{ij} = \frac{\exp(e_{ij})}{\sum_{i=1}^{m}\exp(e_{ij})} \label{eq: att weight softmax}
\end{equation}
With the attention weight, the contextual vector $\boldsymbol{c}_i\in \mathbb{R}^{u}$ for the $i^{th}$ label can be calculated with the hidden vector as follows:
\begin{equation}
    \boldsymbol{c}_i = \sum\limits_{j=1}^{m}\alpha_{ij}\boldsymbol{h}_j \label{eq: context vector c}.
 \end{equation}
Let $P_{y}=[P_{y}^1, P_{y}^2,\cdots,P_{y}^{L_1}]\in \mathbb{R}^{L_1}$ denote the predicted probability for coarse-level labels which can be calculated through a fully connected layer and the Sigmoid function as follows:
\begin{equation}
    P_{y}^i = \frac{1}{1+\exp(\boldsymbol{w}_f\cdot\boldsymbol{c}_i+\boldsymbol{b}_f)}, \label{eq: Sigmoid}
\end{equation}
where $\boldsymbol{w}_f\in \mathbb{R}^{u}$ and $\boldsymbol{b}_f\in \mathbb{R}^{u}$  are the weight and bias vector for the final fully connected layer. Finally, we optimize the binary cross entropy loss between the true coarse label and the predicted probability as follows:
\begin{equation}
    \mathcal{L}_{BCE}(\boldsymbol{y},P_{l_1}) = -\sum\limits_{i=1}^{L_1}[y^i\log(P_{l_1}^i)+ (1-y^i)\log(1-P_{l_1}^i)], \label{eq: BCE_loss}
\end{equation}

% \xinyu{Can we add a paragraph to denote our coarse-level models? Please also provide some equations and notations to be consistent with the later models.}

\subsubsection{Hierarchical attention  \label{sec: proposed Hierarchical attention}}
After getting the embedding representation of the accident reports $\boldsymbol{H}$, we will make the fine-level prediction through the hierarchical attention module. We first calculate the coarse-level relevance score similar to coarse-level attention as follows:
\begin{equation}
    K_{l_1} = V_{l_1}\cdot \boldsymbol{H}^T,
\end{equation}
where $K_{l_1}\in\mathbb{R}^{L_1\times M}$ is the relevance score matrix between the context and coarse labels under a fine-level model. We further collect coarse category distribution $P_{y}\in \mathbb{R}^{L_1}$ from coarse classifier. The coarse category distribution helps us understand the importance of the attention weight from different categories. By applying an element-wise multiplication, we can get the weighted coarse-level relevance score matrix $O_{l_1}\in\mathbb{R}^{L_1\times M}$ as follows:
\begin{equation}
    O_{l_1} = \mathrm{Broadcast}(P_{y}) \otimes  K_{l_1}, 
\end{equation}
where $\mathrm{Broadcast}()$ is an operation to expand $P_y$ so that it has the same dimension as $O_{l_1}$. Our weighted attention matrix $O_{l_1}=(O_{l_1}^1, O_{l_1}^2,\cdots,O_{l_1}^{L_1})\in\mathbb{R}^{L_1\times M}$ comprises $L_1$ vectors where each of them presents the category-wise word relevance within the input documents . In order to get a global attention weight score for each word, We further take a summation over the category dimension and apply the Softmax function to ensure all the computed weights sum to 1 as follows:
\begin{equation}
    W_{l_1} = \mathrm{Softmax}(\sum\limits_{i=1}^{L_1}O_{l_1}^{i}),
\end{equation}
where $ W_{l_1}\in\mathbb{R}^{M}$ represents the global word attention weight for each input instance. 

After calculating the coarse-level attention, we can further develop the fine-level attention weight. Let us first define fine category embedding $V_{l_2}\in \mathbb{R}^{L_2\times u}$ and the attention weight matrix can be calculated as follows: 
\begin{equation}
    W_{l_2} = \mathrm{Softmax}(V_{l_2}\cdot H^T),
\end{equation}
where $W_{l_2}\in\mathbb{R}^{L_2\times M}$ shows the attention weight for each fine-level category. To further incorporate the coarse-level information, we build a linear combination between coarse-level attention and fine-level attention as follows:
\begin{equation}
    W_{l_2}^{'} = W_{l_2} + \lambda \cdot \mathrm{Broadcast}(W_{l_1}),
\end{equation}
where $\lambda$ controls how much we want to leverage the coarse-level guidance. When $\lambda=0$, the formulation becomes a flat classification framework. As we increase $\lambda$, the attention weight will rely more on the coarse level information. 

After getting the updated attention weights, we can further calculate the category-wise contextual representation as follows:
\begin{equation}
    C = W_{l_2}^{'}\cdot H,
\end{equation}
where $C=(C^1, C^2,\dots,C^{L_2})\in\mathbb{R}^{L_2\times u}$ represents $L_2$ contextual representation for each category according its relationship to the input document. 

Then, the fine-level prediction can be achieved through fully connected layers and an activation function as follows: 
\begin{equation}
    P_{l_2} = \sigma(\phi([C \oplus V_{l_2}]\cdot W + \boldsymbol{b})),
\end{equation}
where $W\in\mathbb{R}^{2u\times 1}$ and $\boldsymbol{b}\in\mathbb{R}^{L_2}$ represent the parameters belong to the final fully connected layer. $\oplus$ denotes the concatenation operation to concatenate $C$ and $V_{l_2}$ together. $\phi$ is the non-linear activation function where we apply RELU in our context. 

Finally, the proposed model will optimize the binary cross-entropy loss between the true label distribution and the predicted distribution as follows:
\begin{equation}
    \mathcal{L}_{BCE}(\boldsymbol{z},P_{l_2}) = -\sum\limits_{j=1}^{L_2}[z^j\log(P_{l_2}^j)+ (1-z^j)\log(1-P_{l_2}^j)]. \label{eq: fine BCE}
\end{equation}

\subsubsection{Recursive regularization \label{sec: Recursive regularization proposed}}
The fine-level prediction can be highly improved through the top-down prediction method given coarse-level information. However, as mentioned in the Section~\ref{sec: Recursive Regularization}, we still do not have enough training samples for many fine categories. That means we can not get optimal parameters for the corresponding embedding representation $V_{l_2}$. On the other hand, training samples for the coarse level are usually sufficient. Based on the assumption that the parent category should share similarities with the children category, we can optimize the fine-level embedding by letting it close to its connected coarse-level embedding. We incorporate it into our loss function as a similarity constraint. Let $\boldsymbol{v}_{l_1}^{i}\in\mathbb{R}^u, \forall i=1,\cdots,L_1$ represents the embedding vectors for coarse categories. For $i^{th}$ coarse category, the connected fine-level category embedding are $\boldsymbol{v}_{l_2}^{j}, \forall j\in \pi(i)$. Inspired by the recursive regularization framework in Eq. (\ref{eq: recursive regularization}), the constraint can be expressed as follows:
\begin{equation}
    \mathcal{L}_{similarity} = \sum\limits_{i=1}^{L_1}\sum\limits_{j\in \pi(i)}\frac{1}{2}\|\boldsymbol{v}_{l_1}^{i}-\boldsymbol{v}_{l_2}^{j}\|. \label{eq: proposed similarity}
\end{equation}
\subsubsection{Label distribution penalty  \label{Label distribution penalty-proposed}}
The recursive regularization term helps us build more connections between parent labels and children labels across the label hierarchy. To further improve the rare label identification, we can also use the sibling labels of the rare label. As introduced in Section~\ref{sec: Label Distribution Learning}, a proper label distribution can add additional supervision information to the training process. For those rare labels, we can enrich the label space by leveraging the frequent labels that share similar information to the rare labels. In this work, we propose to use the coarse label distribution to help build fine label distribution according to the strength of the connection between parents and children. Let $P_{l_1}^i,\forall i=1,\cdots,L_1$ be the coarse level probability calculated by the coarse classifiers. For $j^{th}$ fine category, we use $w_j^i, \forall i=1,\cdots,L_1, \forall j=1,\cdots,L_2$ to denote its connection with the $i^{th}$ coarse category. The strength of connection $w_j^i$ is calculated as follows:
\begin{equation}
    w_j^i = \frac{\#(y_i, z_j)}{\#(y_i)},
\end{equation}
where $\#(y_i, z_j)$ represents the number of co-occurrence of coarse category $y_i$ and fine category $z_j$ and $\#(y_i)$ represents number of occurrence of $y_i$. The probability of $j^{th}$ fine category can be calculated as: 
\begin{equation}
    d^j = \frac{\sum\limits_{i=1}^{L_1}P_{l_1}^i w_j^i}{\sum\limits_{i=1}^{L_1}\sum\limits_{j=1}^{L_2}P_{l_1}^i w_j^i}.
\end{equation}
We further use KL divergence to measure the distance between the predicted fine distribution and the expected distribution calculated through coarse distribution as follows:
\begin{equation}
    \mathcal{L}_{distribution}=\sum\limits_{j}(d^{j}\ln\frac{d^{j}}{P_{l_2}^j}). \label{eq: distribution loss}
\end{equation}

\subsubsection{Loss function  \label{sec: loss function}}
After adding the distribution penalty, our final objective becomes:
\begin{equation}
    \mathcal{L} = \mathcal{L}_{BCE} + \lambda_1\mathcal{L}_{similarity} + \lambda_2\mathcal{L}_{distribution}, 
\end{equation}
where $\mathcal{L}_{BCE}$ is the multi-label classification loss defined in Eq. (\ref{eq: fine BCE}), $\mathcal{L}_{similarity}$ is the recursive regularization defined in Eq. (\ref{eq: proposed similarity}), $\mathcal{L}_{distribution}$ is the hierarchical label distribution loss, defined in Eq. (\ref{eq: distribution loss}), and $\lambda_1$ and $\lambda_2$ to control the impact of recursive regularization and label distribution penalty. The best combination of those parameters is found through experiments. When $\lambda_1=0$, the framework becomes a local hierarchical classification approach, and the fine-level embedding vectors are optimized, relying on local information. As $\lambda_1$ increases, the framework becomes a global approach where the fine-level embedding is regularized by the coarse-level embedding. When $\lambda_2=0$, the label space is under a multi-label learning setting. As $\lambda_2$ increases, the framework becomes a label distribution learning setting. It helps enrich the label space and provides additional information for the rare labels. 

\section{Experiment Setup  \label{sec: Experiment Setup}}
In this section, we will introduce more details about the experiment design. An overview of the NTSB dataset will be given in Section \ref{sec: Data Description}. In Section \ref{sec: Implementation details}, we will briefly talk about the implementing platform. We will further introduce the models for comparison and the comparison methods in section \ref{sec: models for comparison} and section \ref{sec: Evaluation Metrics}. After introducing the experiment setup, the major results are discussed in Section \ref{sec: Performance Evaluation} and Section \ref{sec: Rare Category Identification}.

\subsection{Data Description \label{sec: Data Description}}
In this project, we mainly work with the aviation accident reports collected by NTSB. {The NTSB database and its coding system is considered state-of-the-art for aviation safety analysis, given it is the largest and most comprehensive repository of aviation accident data in the United States. This distinction stems not only from its extensive historical coverage, dating back to 1962, but also from its meticulous structuring and coding of a wide array of data points. The database's depth and breadth provide great insights into aviation safety, encompassing various aspects such as aircraft types, operational contexts, environmental conditions, and the sequence of events leading up to accidents. Moreover, the coding system enables detailed and nuanced analysis. It facilitates the identification of trends, patterns, and correlations that might otherwise remain obscured in less comprehensive data sets. This level of detail is crucial for developing targeted safety recommendations, shaping policy, and guiding industry practices. The database's influence extends beyond the United States, often serving as a benchmark for aviation safety standards and practices globally \citep{zhao2022event,zhang2020bayesian,zhang2021sequential, srinivasan2019mining, fuller2020understanding}.}

\begin{figure}[h!]
\centering
\includegraphics[width=0.6\textwidth]{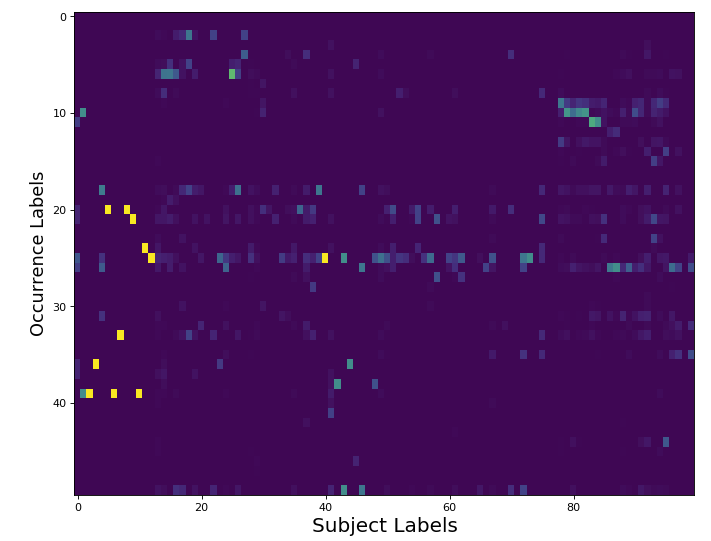}
\caption{Connection matrix between Occurrence labels and Subject labels}
\label{connection}
\end{figure}

From 1982 to 2008, NTSB collects around 62569 accident reports following the same hierarchical labeling strategy. There are 56 Phase codes, 58 Occurrence codes, and 1432 Subject codes \citep{NTSB}. 
The data is publicly accessible and stored in Microsoft Access database format. Figure \ref{distribution} presents the frequency of different labels in the dataset, which suffers a long tail distribution. 
The NTSB database and its coding system is considered the state-of-the-art, given it is the largest 
As shown in the figure, the most frequent subject labels are 19200, 20000, and 20200, with 30886, 24565, and 19231 samples. Those labels represent the most frequent critical events in the aviation accidents which are $\textit{Terrain Condition}$, $\textit{Weather Condition}$, $\textit{Object Condition}$. 250 Subject labels have more than 100 samples, 397 labels have more than 50 samples, and 602 labels have less than 10 samples. We select around $90\%$ accident reports for training and use the rest of those for testing during the modeling process. We further calculate the strength of connection for Occurrence and Subject labels with the training data as mentioned in Section~\ref{Label distribution penalty-proposed}. Figure \ref{connection} gives a brief overview of the connection matrix with the most frequent 100 Subject labels. The X-axis shows the subject labels, and the y-axis shows the occurrence labels. The color of each entry represents the frequency of labels from two levels occurring together. 

\begin{figure}[h!]
\centering
\includegraphics[width=0.6\textwidth]{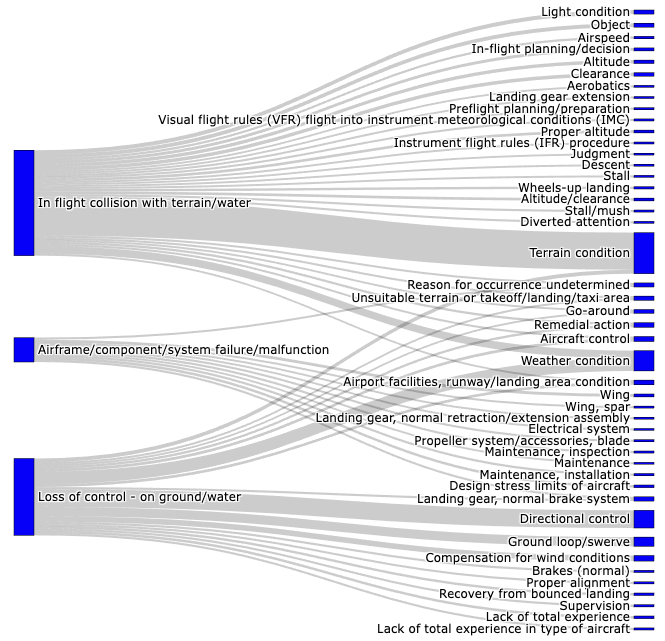}
\caption{An example of label taxonomy defined by NTSB (Airframe malfunction, loss of control and in flight collision)}
\label{taxonomy}
\end{figure}

\subsection{Implementation details \label{sec: Implementation details}}
We apply the BERT model to extract the input documents' semantic representation. The hidden size of the embedding vector is set to 768. To mitigate the overfitting issue, the dropout rate is set to 0.2. We further design the embedding dimension for the hierarchical attention same as the BERT model, considering the consistency.  In order to optimize the parameter, we first limit the range of potential parameters based on previous research from \cite{peng2018large} and \cite{xu2019hierarchical}. We test different $\lambda$ to control the impact from coarse-level attention by splitting the dataset. And we get the best result when $\lambda=2$. For different combination of the three-loss terms $\mathcal{L}_{BCE}$, $\mathcal{L}_{similarity}$ and $\mathcal{L}_{distribution}$, we finally set the corresponding weight to 1, $\lambda_1=0.01$ and $\lambda_2=0.001$. The experiments are all conducted on  Pytorch 1.4.0 on a workstation with Intel(R) Core(TM) i7-5930K CPU @3.50 GHz CPU and accelerated with a single NVIDIA GTX 1080 Ti GPU. 
\subsection{Evaluation Metrics \label{sec: Evaluation Metrics}}
In the hierarchical classification task, we apply the most common measurements used: precision, recall, and F1. Firstly, we will present a global score through micro-averaging to compare the performance of different models. As shown in the following equations, since the micro measurements did not calculate label-wise accuracy, it illustrates the overall accuracy for all samples. 
\begin{align*}
\centering
    P_{micro} &= \frac{\sum_{j=1}^{L}tp_j}{\sum_{j=1}^{L}tp_j+fp_j},\\
    R_{micro} &= \frac{\sum_{j=1}^{L}tp_j}{\sum_{j=1}^{L}tp_j+fn_j},\\
    F_{1-micro} &= \frac{\sum_{j=1}^{L}2tp_j}{\sum_{j=1}^{L}2tp_j+fp_j+fn_j}
\end{align*}
However, micro-f1 tends to favor categories that have large sample sizes when we have an imbalanced dataset. In our problem, the subject labels have a long tail distribution, as shown in Fig. \ref{distribution}. Supervised learning is more capable of learning a good classifier for the frequent labels, which will lead to a pretty high micro-f1 score even though all rare categories are not identified. Thus, we further evaluate the rare category identification using macro-averaging measurements. The macro-averaging calculates the label-wise accuracy and then takes the average of them as follows. 
\begin{align*}
    P_{macro} &= \frac{1}{L}\sum_{j=1}^{L}\frac{tp_j}{tp_j+fp_j},\\
    R_{macro} &= \frac{1}{L}\sum_{j=1}^{L}\frac{tp_j}{tp_j+fn_j},\\
    F_{1-macro} &= \frac{1}{L}\sum_{j=1}^{L}\frac{2tp_j}{2tp_j+fp_j+fn_j}
\end{align*}
To present a fair comparison over frequent categories and rare categories, we further divide all labels into different groups according to their sample size and calculate the group-wise marco averaging. 
\subsection{Models for comparison \label{sec: models for comparison}}
The proposed model is compared with several state-of-the-art models as baselines, which include:
\begin{itemize}
    \item $\textbf{BR-SVM}$ \citep{abedin2010cause}: Under the binary relevance schema, the support vector machine can be treated as a flat classification method for the multi-label task. 
    \item $\bf{SGM}$ \citep{yang2018sgm}: The sequence generation model is another flat classification approach that treats the multi-classification problem as a sequence generation task.  
    \item $\bf{BERT}$ \citep{devlin2018bert}: BERT model provides us a solid baseline to evaluate the effectiveness of the proposed modules. 
\end{itemize}
The variants of the proposed models are listed here:
\begin{itemize}
    \item $\textbf{HABERT}$: Hierarchical attention-based BERT model is the fundamental structure of the proposed model. It leverages the label hierarchy through the hierarchical attention module.
    \item $\textbf{HABERT-R}$: HABERT-R is a variant of HABERT considering the similarity between parent and children nodes through recursive regularization. 
    \item $\textbf{HABERT-L}$: HABERT-L is a variant of HABERT considering the fine category distribution through the coarse category distribution. 
    \item $\textbf{HABERT-RL}$: HABERT-RL is a variant of HABERT using both recursive regularization and label distribution learning penalty. 
\end{itemize}

\begin{table}[h]
\centering\caption{Global comparison for different models with Micro-Precision, Micro-Recall, and Micro-F1. {We conduct T-test of the other benchmarks with the best performers in each category, and we highlight the model with the best performance and the models with similar performance for each metric (not significant under $\alpha=0.05$).}}

\begin{tabular}{|c|ccc|}
\hline
\multirow{2}{*}{Models} & \multicolumn{3}{c|}{Metrics}                                                        \\ \cline{2-4} 
                        & \multicolumn{1}{c|}{Micro-Precision} & \multicolumn{1}{c|}{Micro-Recall} & \ Micro-F1\  \\ \hline
BR-SVM                  & \multicolumn{1}{c|}{0.5356}          & \multicolumn{1}{c|}{0.3386}       & 0.4608   \\ \hline
SGM                     & \multicolumn{1}{c|}{0.5489}          & \multicolumn{1}{c|}{0.4179}       & 0.4745   \\ \hline
BERT                    & \multicolumn{1}{c|}{0.5408}          & \multicolumn{1}{c|}{0.4339}       & 0.5069   \\ \hline
HABERT                  & \multicolumn{1}{c|}{\bf{0.6092}}          & \multicolumn{1}{c|}{0.5083}       &\bf{0.5542}  \\ \hline
HABERT-R                & \multicolumn{1}{c|}{\bf{0.6093}}          & \multicolumn{1}{c|}{0.5082}       & \bf{0.5541} \\ \hline
HABERT-L                & \multicolumn{1}{c|}{0.5743}                & \multicolumn{1}{c|}{\bf{0.5188}}             &0.5452          \\ \hline
HABERT-RL               & \multicolumn{1}{c|}{0.5843}    & \multicolumn{1}{c|}{0.5084}       & 0.5437   \\ \hline
\end{tabular}\label{micro}
\end{table}

\subsection{Performance Evaluation \label{sec: Performance Evaluation}}
This section provides a global evaluation of the proposed methods through micro-averaging. It tells us how different models perform over the whole dataset regardless of the frequency of the categories. Table \ref{micro} presents the results of the experiments. The results show that all the Bert-based models outperform SVM and LSTM models in terms of the F1 score. The original BERT model highly improved the recall rate since the pre-trained model reduces the demand for training samples. We can also validate the efficiency of the hierarchical attention module where HABERT has $9\%$ improvements on f1 over BERT. It made a big improvement over both the precision rate and recall rate. We further show the impact by adding the recursive regularization term. It has been shown that HABERT-R achieves the best precision score and F1 score overall models. It proves the efficiency of adding additional supervision according to the label hierarchy. The f1 score drops a little bit by adding the label distribution penalty. However, the recall rate for both HABERT-L and HABERT-RL increases over other models. The label distribution penalty can help the model identify more rare categories sacrificing some precision accuracy. Overall, the proposed hierarchical attention and recursive regularization can help us get better fine-level classification. Moreover, label distribution learning can further improve rare category classification.

% \usepackage{multirow}
\begin{table}[h!]
\centering\caption{Macro F1 for different models under different sample size. {We conduct T-test of the other benchmarks with the best performers in each category, and we highlight the model with the best performance and the models with similar performance for each metric (not significant under $\alpha=0.05$).}}
\begin{tabular}{|c|cccccc|}
\hline
\multirow{2}{*}{Models} &
  \multicolumn{6}{c|}{Sample Size} \\ \cline{2-7} 
 &
  \multicolumn{1}{c|}{1000-60000} &
  \multicolumn{1}{c|}{500-1000} &
  \multicolumn{1}{c|}{200-500} &
  \multicolumn{1}{c|}{100-200} &
  \multicolumn{1}{c|}{50-100} &
  10-50 \\ \hline
BERT &
  \multicolumn{1}{c|}{0.4356} &
  \multicolumn{1}{c|}{0.3184} &
  \multicolumn{1}{c|}{0.2746} &
  \multicolumn{1}{c|}{0.1723} &
  \multicolumn{1}{c|}{0.1506} &
  0.0368 \\ \hline
HABERT &
  \multicolumn{1}{c|}{0.4709} &
  \multicolumn{1}{c|}{0.3077} &
  \multicolumn{1}{c|}{0.2616} &
  \multicolumn{1}{c|}{0.162} &
  \multicolumn{1}{c|}{0.1373} &
  0.0257 \\ \hline
HABERT-R  & \multicolumn{1}{c|}{0.4752} & \multicolumn{1}{c|}{0.3069} & \multicolumn{1}{c|}{0.2625} & \multicolumn{1}{c|}{0.1592} & \multicolumn{1}{c|}{0.1292} & 0.0233 \\ \hline
HABERT-L &
  \multicolumn{1}{c|}{\bf{0.4881}} &
  \multicolumn{1}{c|}{0.3488} &
  \multicolumn{1}{c|}{0.3154} &
  \multicolumn{1}{c|}{0.1996} &
  \multicolumn{1}{c|}{\bf{0.149}} &
  \bf{0.0221} \\ \hline
HABERT-RL & \multicolumn{1}{c|}{0.4796} & \multicolumn{1}{c|}{\bf{0.3575}} & \multicolumn{1}{c|}{\bf{0.3264}} & \multicolumn{1}{c|}{\bf{0.2144}} & \multicolumn{1}{c|}{\bf{0.1515}} & \bf{0.0243} \\ \hline
\end{tabular}\label{macro-f1}
\end{table}

% \usepackage{multirow}
% \usepackage{multirow}

\begin{table}[h!]
\centering\caption{Macro Precision for different models under different sample size. {We conduct T-test of the other benchmarks with the best performers in each category, and we highlight the model with the best performance and the models with similar performance for each metric (not significant under $\alpha=0.05$).}}
\begin{tabular}{|c|cccccc|}
\hline
\multirow{2}{*}{Models} &
  \multicolumn{6}{c|}{Sample Size} \\ \cline{2-7} 
 &
  \multicolumn{1}{c|}{1000-60000} &
  \multicolumn{1}{c|}{500-1000} &
  \multicolumn{1}{c|}{200-500} &
  \multicolumn{1}{c|}{100-200} &
  \multicolumn{1}{c|}{50-100} &
  10-50 \\ \hline
BERT &
  \multicolumn{1}{c|}{0.4916} &
  \multicolumn{1}{c|}{0.3923} &
  \multicolumn{1}{c|}{0.3282} &
  \multicolumn{1}{c|}{0.208} &
  \multicolumn{1}{c|}{0.1782} &
  0.0431 \\ \hline
HABERT &
  \multicolumn{1}{c|}{\bf{0.5034}} &
  \multicolumn{1}{c|}{0.3642} &
  \multicolumn{1}{c|}{0.2869} &
  \multicolumn{1}{c|}{0.1774} &
  \multicolumn{1}{c|}{0.1591} &
  0.0269 \\ \hline
HABERT-R &
  \multicolumn{1}{c|}{0.4942} &
  \multicolumn{1}{c|}{0.3558} &
  \multicolumn{1}{c|}{0.3108} &
  \multicolumn{1}{c|}{0.19} &
  \multicolumn{1}{c|}{0.1543} &
  0.0251 \\ \hline
HABERT-L &
  \multicolumn{1}{c|}{0.4869} &
  \multicolumn{1}{c|}{\bf{0.3979}} &
  \multicolumn{1}{c|}{0.3713} &
  \multicolumn{1}{c|}{0.2362} &
  \multicolumn{1}{c|}{\bf{0.1889}} &
  0.0257 \\ \hline
HABERT-RL &
  \multicolumn{1}{c|}{0.4905} &
  \multicolumn{1}{c|}{\bf{0.4002}} &
  \multicolumn{1}{c|}{\bf{0.3794}} &
  \multicolumn{1}{c|}{\bf{0.2815}} &
  \multicolumn{1}{c|}{\bf{0.1906}} &
  \bf{0.0317} \\ \hline
\end{tabular}\label{macro-precision}
\end{table}

% \usepackage{multirow}
\begin{table}[h]
\centering\caption{Macro recall for different models under different sample size. {We conduct T-test of the other benchmarks with the best performers in each category, and we highlight the model with the best performance and the models with similar performance for each metric (not significant under $\alpha=0.05$).}}
\begin{tabular}{|c|cccccc|}
\hline
\multirow{2}{*}{Models} &
  \multicolumn{6}{c|}{Sample Size} \\ \cline{2-7} 
 &
  \multicolumn{1}{c|}{1000-60000} &
  \multicolumn{1}{c|}{500-1000} &
  \multicolumn{1}{c|}{200-500} &
  \multicolumn{1}{c|}{100-200} &
  \multicolumn{1}{c|}{50-100} &
  10-50 \\ \hline
BERT &
  \multicolumn{1}{c|}{0.4443} &
  \multicolumn{1}{c|}{0.3157} &
  \multicolumn{1}{c|}{0.2868} &
  \multicolumn{1}{c|}{0.1797} &
  \multicolumn{1}{c|}{0.1522} &
  0.0378 \\ \hline
HABERT &
  \multicolumn{1}{c|}{0.4844} &
  \multicolumn{1}{c|}{0.3021} &
  \multicolumn{1}{c|}{0.2761} &
  \multicolumn{1}{c|}{0.1803} &
  \multicolumn{1}{c|}{0.142} &
  0.0289 \\ \hline
HABERT-R &
  \multicolumn{1}{c|}{0.4955} &
  \multicolumn{1}{c|}{0.299} &
  \multicolumn{1}{c|}{0.2616} &
  \multicolumn{1}{c|}{0.157} &
  \multicolumn{1}{c|}{0.1262} &
  0.0246 \\ \hline
HABERT-L &
  \multicolumn{1}{c|}{\bf{0.5137}} &
  \multicolumn{1}{c|}{\bf{0.3382}} &
  \multicolumn{1}{c|}{\bf{0.3094}} &
  \multicolumn{1}{c|}{\bf{0.1969}} &
  \multicolumn{1}{c|}{\bf{0.1379}} &
  \bf{0.0211} \\ \hline
HABERT-RL &
  \multicolumn{1}{c|}{0.4787} &
  \multicolumn{1}{c|}{\bf{0.3398}} &
  \multicolumn{1}{c|}{\bf{0.3113}} &
  \multicolumn{1}{c|}{\bf{0.197}} &
  \multicolumn{1}{c|}{\bf{0.1376}} &
  \bf{0.0217} \\ \hline
\end{tabular}\label{macro-recall}
\end{table}

% \usepackage{multirow}
\begin{table}[h!]
\centering\caption{Category-wise f1 score comparison under few-shot learning setting}
\hspace*{-1cm}
\resizebox{0.75\columnwidth}{!}{%
\begin{tabular}{|c|ccccc|}
\hline
\multirow{2}{*}{Category} &
  \multicolumn{5}{c|}{Model} \\ \cline{2-6} 
 &
  \multicolumn{1}{c|}{BERT} &
  \multicolumn{1}{c|}{HABERT} &
  \multicolumn{1}{c|}{HABERT-R} &
  \multicolumn{1}{c|}{HABERT-L} &
  HABERT-RL \\ \hline
IDENTIFICATION OF AIRCRAFT ON RADAR (28) &
  \multicolumn{1}{c|}{0} &
  \multicolumn{1}{c|}{0} &
  \multicolumn{1}{c|}{0} &
  \multicolumn{1}{c|}{0} &
  \bf{0.381} \\ \hline
EXHAUST SYSTEM, CLAMP (39) &
  \multicolumn{1}{c|}{0.1053} &
  \multicolumn{1}{c|}{0.1053} &
  \multicolumn{1}{c|}{\bf{0.1818}} &
  \multicolumn{1}{c|}{0.1333} &
  0 \\ \hline
TRAFFIC ADVISORY (137) &
  \multicolumn{1}{c|}{0} &
  \multicolumn{1}{c|}{0} &
  \multicolumn{1}{c|}{0} &
  \multicolumn{1}{c|}{0.0377} &
  \bf{0.2738} \\ \hline
WAKE TURBULENCE (99) &
  \multicolumn{1}{c|}{0.2513} &
  \multicolumn{1}{c|}{0} &
  \multicolumn{1}{c|}{\bf{0.3288}} &
  \multicolumn{1}{c|}{0.1311} &
  0.3892 \\ \hline
LUBRICATING SYSTEM, OIL FILTER/SCREEN (50) &
  \multicolumn{1}{c|}{\bf{0.1778}} &
  \multicolumn{1}{c|}{0.1176} &
  \multicolumn{1}{c|}{0.069} &
  \multicolumn{1}{c|}{0.1176} &
  0 \\ \hline
MAINTENANCE, LUBRICATION (70) &
  \multicolumn{1}{c|}{0.0408} &
  \multicolumn{1}{c|}{0} &
  \multicolumn{1}{c|}{0} &
  \multicolumn{1}{c|}{0.069} &
  \bf{0.2069} \\ \hline
VACUUM SYSTEM (53) &
  \multicolumn{1}{c|}{0.2581} &
  \multicolumn{1}{c|}{0.1053} &
  \multicolumn{1}{c|}{0.1818} &
  \multicolumn{1}{c|}{0.1818} &
  \bf{0.381} \\ \hline
ELEVATOR TRIM (72) &
  \multicolumn{1}{c|}{0.1488} &
  \multicolumn{1}{c|}{0.1343} &
  \multicolumn{1}{c|}{0.2687} &
  \multicolumn{1}{c|}{0.2319} &
  \bf{0.2703} \\ \hline
LOSS OF TAIL ROTOR EFFECTIVENESS (85) &
  \multicolumn{1}{c|}{\bf{0.252}} &
  \multicolumn{1}{c|}{0.1356} &
  \multicolumn{1}{c|}{0.1039} &
  \multicolumn{1}{c|}{0.1356} &
  0.16 \\ \hline
MIXTURE CONTROL, CABLE (53) &
  \multicolumn{1}{c|}{0.2717} &
  \multicolumn{1}{c|}{0.2078} &
  \multicolumn{1}{c|}{0.289} &
  \multicolumn{1}{c|}{0.3077} &
  \bf{0.3876} \\ \hline
ENG ASSEMBLY, BLOWER /IMPELLER/INTEGRAL (77) &
  \multicolumn{1}{c|}{0.1835} &
  \multicolumn{1}{c|}{0.2509} &
  \multicolumn{1}{c|}{0.3216} &
  \multicolumn{1}{c|}{0.3368} &
  \bf{0.3821} \\ \hline
REMOVAL OF CONTROL/GUST LOCK (39) &
  \multicolumn{1}{c|}{0.2278} &
  \multicolumn{1}{c|}{0.2278} &
  \multicolumn{1}{c|}{\bf{0.2535}} &
  \multicolumn{1}{c|}{0.2069} &
  0.2105 \\ \hline
FLT CONTROL SYST, WING SPOILER SYSTEM (42) &
  \multicolumn{1}{c|}{0.2813} &
  \multicolumn{1}{c|}{0.2951} &
  \multicolumn{1}{c|}{\bf{0.45}} &
  \multicolumn{1}{c|}{0.3077} &
  0.1429 \\ \hline
ICE/FROST REMOVAL FROM AIRCRAFT (133) &
  \multicolumn{1}{c|}{0.2045} &
  \multicolumn{1}{c|}{0.1778} &
  \multicolumn{1}{c|}{0.1509} &
  \multicolumn{1}{c|}{0.2046} &
  \bf{0.3273} \\ \hline
FUSELAGE, SEAT (86) &
  \multicolumn{1}{c|}{0.3173} &
  \multicolumn{1}{c|}{0.2793} &
  \multicolumn{1}{c|}{0.252} &
  \multicolumn{1}{c|}{0.3404} &
  \bf{0.4673} \\ \hline
LANDING GEAR, STEERING SYSTEM (128) &
  \multicolumn{1}{c|}{0.1928} &
  \multicolumn{1}{c|}{0.1667} &
  \multicolumn{1}{c|}{0.0202} &
  \multicolumn{1}{c|}{0.2792} &
  \bf{0.3077} \\ \hline
\end{tabular}\label{category}
}
\end{table}

\begin{figure}[h!]
\centering
\includegraphics[width=1\textwidth]{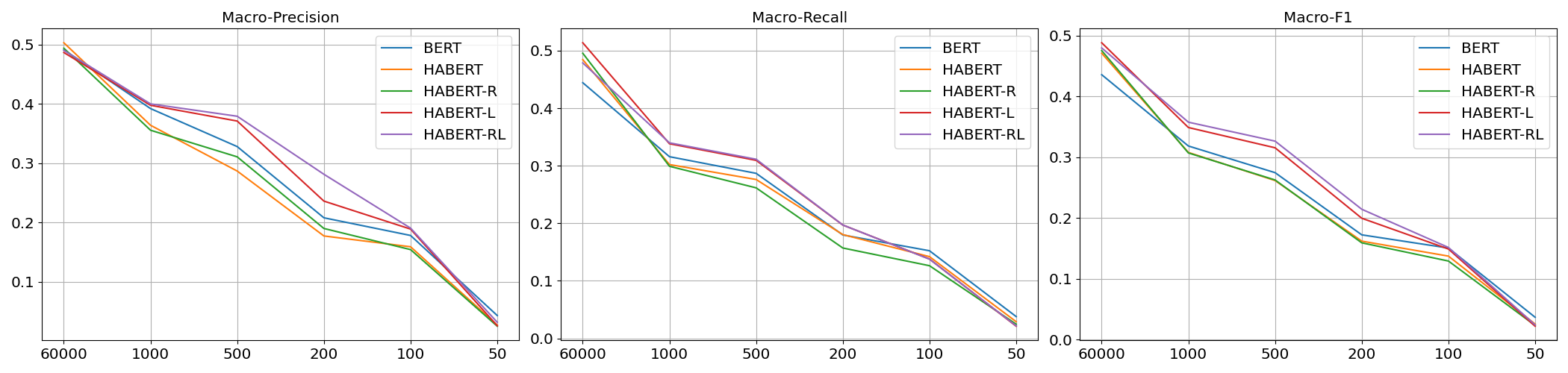}
\caption{Model comparison under different sample size)}
\label{rare}
\end{figure}
\subsection{Rare Category Identification \label{sec: Rare Category Identification}}
We further provide more experiment results to discuss the influence of sample size on the model performance. To ensure all categories have the same contribution, we will apply macro-averaging metrics in this section. We divide all data into 6 groups according to their sample size. After the processing, 46 categories have sample sizes from 1000 to 60000, and 43 categories have sample sizes from 500 to 1000, which can provide sufficient samples for training. 69 categories have sample sizes within 200 - 500, 92 categories have sample sizes within 100 - 200, and 146 categories have sample sizes within 50 - 100. Those categories provide fewer training samples that belong to the few-shot learning settings. Other categories with less than 50 samples are usually very challenging for deep learning methods. 399 categories have sample sizes within 10 - 50, and 602 categories have sample sizes less than 10.

Figure \ref{rare} presents macro-level comparison for different models under different sample size. We directly take the average of precision, recall, and f1 for categories belonging to each group. The x-axis shows the upper bound for the corresponding group, and the y-axis shows the metric's value. The figure shows that the performance from all methods decreases as the sample size decreases. The proposed HABERT-RL can improve the HABERT for all groups in terms of macro-f1. We can get a more accurate evaluation through Table \ref{macro-f1}, Table \ref{macro-recall} and Table \ref{macro-precision}. For frequent categories, HABERT-RL has made $1.8\%$ improvements over HABERT when the sample size is larger than 1000 in terms of macro-f1. Larger improvements are made when the sample size decreases. $16.2\%$ improvement is made for groups 500 -1000. $24.8\%$ improvement is made for groups 200 -500. $32.3\%$ improvement is made for groups 100-200. For categories with less than 100 samples, none of the models can perform well, and the improvements are not very significant. $10.3\%$ improvement is made for group 50 -100. None improvement is made for group 10-50. The most significant improvement our model made is within the categories with a small sample size from 50 - 200. It proves the effectiveness of the label distribution penalty. 

We further present the F1 score for selected categories under a few-shot learning setting. As shown in Table \ref{category}, the left column is the names of each category, and the corresponding sample size for each category is included in the parentheses. Those are categories that have sample sizes between 10 and 200. By applying the proposed penalty terms, HABERT-RL can achieve significant improvements over other methods. For rare categories such as IDENTIFICATION OF AIRCRAFT ON RADAR and TRAFFIC ADVISORY, only HABERT-RL is able to identify those from the reports. For categories that have slightly more samples, such as ICE/FROST REMOVAL FROM AIRCRAFT and LANDING GEAR STEERING SYSTEM, HABERT-RL can also make further improvements over other methods. 

% \xinyu{May consider show some examples of attention module in fine or coarse levels}

\section{Conclusion  \label{sec: Conclusion}}
This work aims to discuss the feasibility of building an information extraction system for fine-level events in the aviation domain. Our work leverages the event taxonomy defined by the domain expert from NTSB and converts the problem into a hierarchical classification task. Since fine-level events have a very large label size, the long-tail distribution of the data leads to the biggest challenge for implementing an accurate algorithm. To tackle this challenge, we propose to extend the state-of-the-art BERT model with a novel multi-label hierarchical classification model. First, we develop a hierarchical attention module to introduce coarse-level information into the fine category classification process. Our experiment validates the efficiency of this module, where we achieve $20\%$ improvements over widely used BR-SVM in terms of micro-F1. The second component is to provide additional supervision to the fine-level parameters through recursive regularization. The experiment shows that the proposed HABERT-R achieves the best performance among all models. Finally, we discuss how to improve the classification accuracy for categories with small training samples. We propose a label distribution penalty term in the model and evaluate the rare category identification through the macro-F1 score. Our results show that the label distribution penalty can significantly improve the rare category classification accuracy. $32.3\%$ improvement is made when the sample size is between 100 and 200. In summary, we are the first work discussing the possibility of extracting fine-level information in the aviation domain for the NTSB dataset.  

\section*{Acknowledgments}
The  research reported in this paper was supported by funds from NASA University Leadership Initiative program (Contract No. NNX17AJ86A, Project Officer: Dr. Anupa Bajwa, Principal Investigator: Dr. Yongming Liu). {The support is gratefully acknowledged. We would also thank the two experts in the aviation accident report and safety analysis, namely, Dan Larseon an aviation analytics from Metron Aviation, and Xue Ping, a computational linguist from Boeing (Retired). They have worked with the team on supervising the research development and help the team understand the current NTSB coding system. }

% References here (outcomment the appropriate case)

% CASE 1: BiBTeX used to constantly update the references
%   (while the paper is being written).
%\bibliographystyle{informs2014} % outcomment this and next line in Case 1
%\bibliography{<your bib file(s)>} % if more than one, comma separated

% CASE 2: BiBTeX used to generate mypaper.bbl (to be further fine tuned)
%\input{mypaper.bbl} % outcomment this line in Case 2

%If you don't use BiBTex, you can manually itemize references as shown below.

% \bibliographystyle{nonumber}
% \bibliography{sample}
\begin{small}
    \bibliography{sample.bib}
    \bibliographystyle{informs2014}
\end{small}

%%%%%%%%%%%%%%%%%
\end{document}